\documentclass[letterpaper]{article} 
\usepackage[preprint]{aaai2027}  
\usepackage[hyphens]{url}  
\usepackage{graphicx} 
\usepackage{natbib}  
\usepackage{caption} 
\usepackage{algorithm}
\usepackage{algorithmic}

\usepackage{newfloat}
\usepackage{tabularx}
\usepackage{amsfonts}
\usepackage{amsmath}
\usepackage{listings}
\usepackage{multirow}
\DeclareCaptionStyle{ruled}{labelfont=normalfont,labelsep=colon,strut=off} 
\floatstyle{ruled}
\newfloat{listing}{tb}{lst}{}
\floatname{listing}{Listing}
\usepackage[table]{xcolor}
\definecolor{rowgray}{gray}{0.92}
\usepackage{booktabs}

\title{SPARGen: Unifying Spatial Perception and Reasoning \\
through Native Multimodal Generation}
\author{
    Jinsheng Quan\textsuperscript{\rm 1,2}, 
    Jianhua Li\textsuperscript{\rm 2}, 
    Siyi Xie\textsuperscript{\rm 2,3}, 
    Xuanke Shi\textsuperscript{\rm 2}, 
    Kewang Deng\textsuperscript{\rm 2}, 
    Zukai Chen\textsuperscript{\rm 2}, 
    Feifei Shao\textsuperscript{\rm 1}, 
    Lei Yang\textsuperscript{\rm 2}, 
    Quan Wang\textsuperscript{\rm 2}\corresponding, 
    Yawei Luo\textsuperscript{\rm 1}\corresponding
}
\affiliations{
    \textsuperscript{\rm 1}Zhejiang University
    \textsuperscript{\rm 2}SenseTime Research
    \textsuperscript{\rm 3}Peking University
}

\makeatletter
\let\original@maketitle\@maketitle
\def\@maketitle{%
    \original@maketitle
    \begingroup
        \centering
        \includegraphics[width=\textwidth]{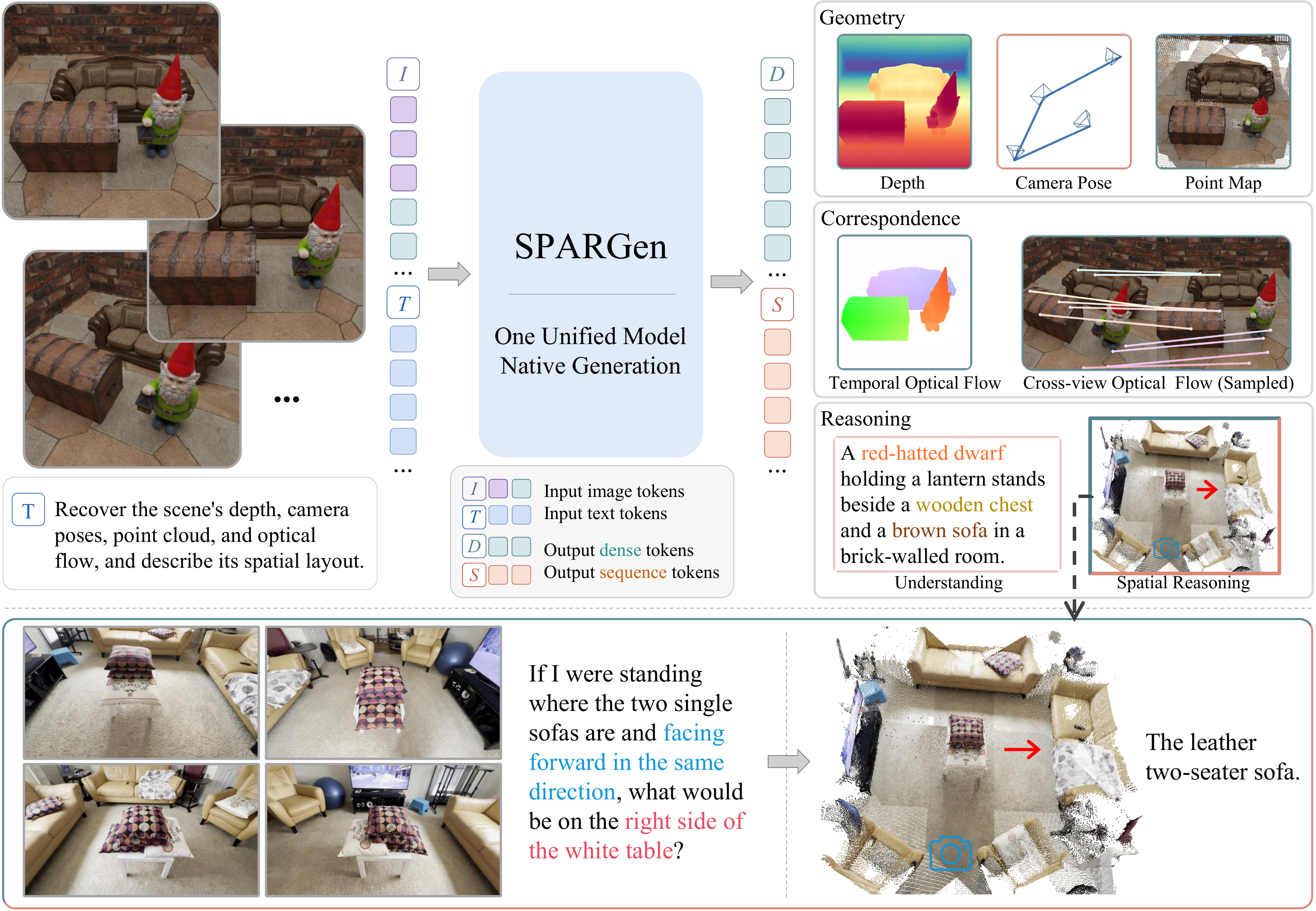}\par
        \captionof{figure}{\textbf{SPARGen unifies spatial perception and reasoning through native multimodal generation.} Conditioned on visual observations and natural-language instructions, a single model generates dense geometric fields and token sequences for 3D reconstruction, correspondence estimation, visual understanding, and spatial reasoning, without task-specific modules.}
        \label{fig:pre}
    \endgroup
}
\makeatother

\begin{document}

\maketitle

\begin{abstract}
Spatial perception and reasoning from visual observations require recovering geometric structure, establishing correspondences, and understanding spatial relations. Existing approaches typically address these capabilities separately using task-specific architectures or external geometric modules, limiting knowledge transfer among complementary representations of the same physical scene. We introduce SPARGen, a unified multimodal framework that casts 3D reconstruction, dense correspondence, and spatial reasoning as instruction-conditioned generation tasks. SPARGen serializes compact structured and linguistic outputs as token sequences while generating dense geometric fields in image-aligned forms, enabling spatial supervision to jointly shape shared representations within a native multimodal generative model. Experiments across benchmarks for 3D reconstruction, correspondence, and spatial reasoning show that SPARGen achieves competitive performance across heterogeneous spatial tasks within a single native multimodal generative framework.
\end{abstract}


\section{Introduction}
Recovering the geometry of a 3D scene and reasoning about it are complementary capabilities for perceiving and interacting with the physical world. Recent advances have made substantial progress in geometric reconstruction~\cite{wang2024vggsfm,wang2025vggt,lin2025depth}, correspondence estimation~\cite{zhangufm,edstedt2024roma,huang2022flowformer}, and spatial reasoning~\cite{chen2024spatialvlm,azuma2022scanqa,ma2025spatialllm}. These capabilities play different but interconnected roles: reconstruction recovers scene structure, correspondence associates evidence across views or time, and reasoning converts spatial representations into task-relevant conclusions.

Unifying these capabilities is desirable because they provide complementary supervision for the same underlying physical scene. Depth, camera motion, point maps, and optical flow jointly characterize scene structure and its variation across observations, while spatial question answering requires abstracting such geometric evidence into semantic relations. Learning these tasks together therefore has the potential to promote knowledge transfer across tasks, improve the consistency of spatial predictions, and connect geometric scene understanding with high-level reasoning. 

However, existing approaches address only parts of this problem. Feed-forward geometry models provide accurate depth, point maps, camera poses, or correspondences, but generally do not support spatial question answering through the same model~\cite{lin2025depth,wang2025continuous,wang2025pi,cong2026flow3r}. Multimodal models offer flexible instruction following and spatial reasoning, yet typically receive limited supervision from dense geometry and correspondences~\cite{huang2024chat,chen2024grounded,deng20253d}. Recent unified models combine geometric perception with multimodal understanding, but often introduce geometry-specific encoders, regression heads, or external modules~\cite{qi2024gpt4point,hu2026g,xu2025uniugg}. Consequently, existing systems provide limited opportunities for dense geometric, correspondence, and semantic supervision to jointly shape shared spatial representations.

In this paper, we present SPARGen, a unified framework for spatial perception and reasoning, as illustrated in Fig.~\ref{fig:pre}. Our design is motivated by a simple view: spatial intelligence is not merely a collection of isolated tasks, but a process of constructing spatial representations from observations and using them to answer conditioned queries. Accordingly, SPARGen formulates geometric reconstruction, correspondence estimation, and spatial reasoning as instruction-conditioned generation tasks within a single multimodal model. Built on Bagel~\cite{deng2025emerging}, SPARGen predicts depth maps, camera poses, and point maps for 3D reconstruction; optical flow for dense correspondence estimation; and textual answers for spatial question answering.

To accommodate these heterogeneous outputs, SPARGen adopts two complementary generative formulations. Dense spatial fields, including depth maps, point maps, and optical flow, are encoded as image-like representations and generated through the model's native image-generation pathway. Compact structured outputs, such as camera poses, as well as answers to spatial questions, are serialized as token sequences and generated through the autoregressive pathway. Both generative pathways are integrated within a shared Mixture-of-Transformer-Experts (MoT) backbone, allowing geometric, correspondence, and semantic supervision to jointly shape the model's multimodal representations. SPARGen thus provides a shared instruction interface for constructing and using spatial representations without introducing task-specific regression heads or external geometric prediction modules.

This native generative formulation distinguishes SPARGen by how spatial capabilities are incorporated into a foundation model. Prior approaches typically specialize in either dense geometric prediction or language-based spatial reasoning~\cite{wang2025vggt,lin2025depth,chen2024spatialvlm,wu2026spatial}. SPARGen instead supports both within a single instruction-conditioned multimodal architecture. More closely related to our work, G$^2$VLM~\cite{hu2026g} unifies language understanding and geometric prediction by introducing geometry-specific components. In contrast, SPARGen starts from a pretrained unified multimodal model and represents heterogeneous spatial targets in forms that can be directly produced by its native image-generation and autoregressive pathways. This design enables dense geometric fields, dense correspondences, and structured spatial answers to be learned through a shared generative interface, without task-specific prediction heads or external geometric modules. Our main contributions are as follows:
\begin{itemize}
    \item We formulate 3D reconstruction, dense correspondence, and spatial reasoning as instruction-conditioned generation tasks, representing dense outputs as image-aligned fields and structured outputs as token sequences.
    
    \item We instantiate this formulation in SPARGen, which leverages the native image-generation and autoregressive pathways of a shared MoT backbone to support heterogeneous spatial tasks without external geometric modules.

    \item Experiments demonstrate that a single SPARGen model achieves competitive performance across visual-geometry, optical flow, and spatial-reasoning benchmarks. Ablations further reveal positive interactions among heterogeneous supervision.
\end{itemize}

\begin{figure*}
    \centering
    \includegraphics[width=1.0\linewidth]{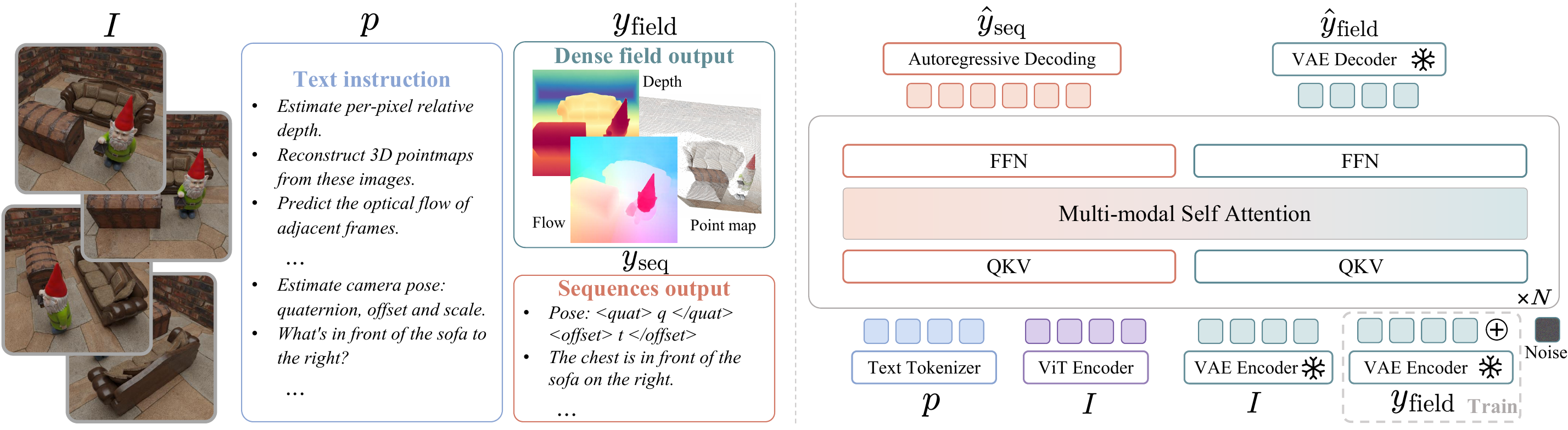}
        \caption{\textbf{Overview of SPARGen.} Given a sequence of RGB images and a language instruction, SPARGen represents spatial outputs as sequences or image-aligned fields. These two output formats are generated through the native autoregressive and rectified-flow pathways of a shared MoT backbone. The dashed box denotes the training-only encoding of target fields.} 
    \label{fig:overview}
    \vspace{-0.2cm}
\end{figure*}

\section{Related Work}
\textbf{3D Reconstruction and Dense Correspondence.} 
Visual geometry systems recover camera motion and scene structure~\cite{schonberger2016sfm}. More recently, feed-forward reconstruction models have moved toward broader geometric representations. DUSt3R casts uncalibrated stereo reconstruction as pointmap regression~\cite{wang2024dust3r,Yang2025Fast3R}. VGGT jointly predicts camera poses, depth, point maps, and tracks from one or multiple views~\cite{wang2025vggt}, while VGGT-$\Omega$ scales this paradigm to substantially larger models and datasets and extends it to dynamic scenes~\cite{wang2026vggtomega}. A complementary line of work repurposes generative priors for dense prediction: Marigold adapts a diffusion model for depth estimation~\cite{ke2024marigold}, and Edit2Perceive uses an image-editing model for depth estimation, surface-normal estimation, and matting~\cite{shi2026edit2perceive}. Similarly, learning-based matching methods replace individual stages of traditional correspondence pipelines: LoFTR performs detector-free matching with transformers~\cite{sun2021loftr}, while DKM~\cite{edstedt2023dkm} and RoMa~\cite{edstedt2024roma} directly predict dense correspondences for robust two-view geometry. These methods demonstrate increasingly general geometric perception, but typically expose their capabilities through fixed pipelines and geometric objectives. In contrast, SPARGen casts both reconstruction and correspondence estimation as instruction-conditioned behaviors of a multimodal generative model.

\textbf{Geometry-Aware Spatial Reasoning.} 
Vision-language models acquire strong semantic knowledge but remain unreliable in recovering geometric structure. SpatialVLM addresses this limitation by training on large-scale spatial question-answering data~\cite{chen2024spatialvlm}, whereas SpatialRGPT combines region-level supervision with a plug-in depth representation~\cite{cheng2024spatialrgpt}. Other approaches inject stronger reconstruction priors. VLM-3R derives implicit spatial and camera tokens from a geometry encoder for monocular video reasoning~\cite{fan2026vlm}, and Spatial-MLLM combines semantic and geometry-oriented visual encoders~\cite{ma2025spatialllm}. More tightly coupled designs learn geometry within the multimodal model: G$^2$VLM adopts dedicated geometric and semantic transformer experts with shared attention to support reconstruction and spatial reasoning~\cite{hu2026g}. Collectively, these works establish that explicit geometry benefits spatial reasoning. Nevertheless, geometric information is commonly provided by an external encoder or modeled through dedicated experts or output heads, and most systems combine language reasoning with only a limited set of geometric capabilities. SPARGen instead places spatial perception and reasoning under the same instruction-conditioned training interface.

\textbf{Unified Multimodal Understanding and Generation.} 
Recent multimodal foundation models seek to replace separate understanding and generation systems with a shared model. Chameleon represents images and text as interleaved discrete tokens in an early-fusion architecture~\cite{chameleon2024}; Show-o combines autoregressive language modeling with discrete diffusion in a single transformer~\cite{xie2025showo}; and Janus decouples the visual encoding pathways for understanding and generation while retaining a shared autoregressive backbone~\cite{wu2025janus}. Bagel further scales unified pretraining on interleaved text and image data, exhibiting broad multimodal reasoning and generation capabilities~\cite{deng2025emerging}. These models are primarily developed for semantic understanding, content generation, and image editing, leaving precise geometric prediction and dense correspondence comparatively underexplored.

\section{Method}
\label{sec:method}

We first introduce the problem formulation and provide an overview of SPARGen in Sec.~\ref{sec:overview}. We then describe its architecture and two native output representations in Sec.~\ref{sec:output_representation}, followed by the corresponding training objectives in Sec.~\ref{sec:training_objectives}.

\subsection{Problem Formulation and Overview}
\label{sec:overview}
Given a sequence of RGB images $\mathcal{I}=\{I_i\}_{i=1}^{N}$, where $I_i\in\mathbb{R}^{H\times W\times 3}$, and a tokenized natural-language instruction $p=(p_l)_{l=1}^{L_p}\in\mathcal{V}^{L_p}$, our goal is to predict a target $Y_\tau$ for a spatial task $\tau\in\mathcal{T}=\mathcal{T}_{\mathrm{seq}}\cup\mathcal{T}_{\mathrm{field}}$. Here, $\mathcal{V}$ denotes the tokenizer vocabulary, and $\mathcal{T}_{\mathrm{seq}}$ and $\mathcal{T}_{\mathrm{field}}$ denote the sets of token-sequence and dense-field generation tasks, respectively.

Rather than introducing a separate module for each task, SPARGen represents spatial targets using two native generative formats:
\begin{equation}
\mathcal{R}_{\tau}(Y_\tau)=
\begin{cases}
    \mathcal{S}_{\tau}(Y_\tau)
    \in \mathcal{V}^{L_\tau},
    & \tau\in\mathcal{T}_{\mathrm{seq}}, \\[2mm]
    \Phi_{\tau}(Y_\tau)
    \in
    \mathbb{R}^{M_\tau\times H\times W\times 3},
    & \tau\in\mathcal{T}_{\mathrm{field}}.
\end{cases}
\label{eq:unified_representation}
\end{equation}
Here, $\mathcal{S}_{\tau}$ serializes a sequence target into $L_\tau$ discrete tokens, whereas $\Phi_{\tau}$ encodes a dense target as $M_\tau$ image-aligned fields. The conditional generative model is defined as:
\begin{equation}
p_{\theta}(
    \mathcal{R}_{\tau}(Y_\tau)
    \mid
    \mathcal{I}, p
).
\label{eq:conditional_generation}
\end{equation}
This formulation unifies diverse spatial tasks while preserving the structural properties of their respective outputs.

As illustrated in Fig.~\ref{fig:overview}, SPARGen builds on the MoT architecture of Bagel~\cite{deng2025emerging}. The input images are encoded into visual-understanding tokens by a ViT encoder, while the instruction is represented as text tokens. Text and visual tokens interact through joint multimodal self-attention. Sequence targets are generated autoregressively, whereas dense fields are generated in the VAE latent space through rectified flow. During training, the frozen VAE encoder maps the target fields to clean target latents, which are interpolated with Gaussian noise to construct noisy visual tokens for rectified-flow training.

\subsection{Unified Generative Architecture}
\label{sec:output_representation}

\paragraph{Unified Architecture for Spatial Generation.} 
SPARGen adopts the MoT architecture, which consists of an understanding expert and a generation expert. The two token streams use modality-specific projections and feed-forward networks but interact through joint multimodal self-attention. This design allows semantic and geometric information to interact throughout the backbone while retaining their respective generation mechanisms.

Given an image sequence $\mathcal{I}$ and an instruction $p$, the conditioning contexts for token-sequence and dense-field generation are constructed as
\begin{equation}
\begin{aligned}
C_{\mathrm{seq}}
&=\operatorname{Concat}(
    E_{\mathrm{text}}(p),
    E_{\mathrm{vit}}(\mathcal{I})),\\
C_{\mathrm{field}}
&=\operatorname{Concat}(
    E_{\mathrm{text}}(p),
    E_{\mathrm{vit}}(\mathcal{I}),
    E_{\mathrm{vae}}(\mathcal{I})),
\end{aligned}
\label{eq:multimodal_context}
\end{equation}
where $E_{\mathrm{text}}$ denotes the text embedding module,
$E_{\mathrm{vit}}$ denotes the visual-understanding encoder, and
$E_{\mathrm{vae}}$ denotes the frozen VAE encoder. Thus, dense-field generation is conditioned on both the ViT and VAE tokens of the input images.

For a token-sequence task $\tau\in\mathcal{T}_{\mathrm{seq}}$, let $\mathcal{S}_{\tau}(Y_\tau)=(s_1,\ldots,s_{L_\tau})$ denote its serialized target. The model generates the sequence autoregressively:
\begin{equation}
p_{\theta}(
    \mathcal{S}_{\tau}(Y_\tau)
    \mid \mathcal{I},p
)
=
\prod_{j=1}^{L_\tau}
p_{\theta}(s_j\mid C_{\mathrm{seq}},s_{<j}).
\label{eq:sequence_generation}
\end{equation}

For a dense-field task $\tau\in\mathcal{T}_{\mathrm{field}}$, the target is first mapped to a clean latent by the frozen VAE encoder $z_0=E_{\mathrm{vae}}(\Phi_{\tau}(Y_\tau))$. We then construct a linear rectified-flow path between the clean latent $z_0$ and Gaussian noise $\epsilon$:
\begin{equation}
z_t=(1-t)z_0+t\epsilon,
\quad
t\sim\mathcal{U}(0,1),
\quad
\epsilon\sim\mathcal{N}(0,\mathbf{I}).
\label{eq:flow_path}
\end{equation}
The target velocity is $u_t=\epsilon-z_0$. Conditioned on the multimodal context, the generation pathway predicts the velocity field $\widehat{v}_t=v_{\theta}(z_t,t;C_{\mathrm{field}})$, which is trained to match $u_t$.

At inference time, sampling starts from $z_1\sim\mathcal{N}(0,\mathbf{I})$ and integrates the predicted velocity field from $t=1$ to $t=0$. The resulting latent $\widehat{z}_0$ is decoded by the frozen VAE decoder.

\paragraph{Sparse and Structured Outputs as Sequences.}
For each task $\tau\in\mathcal{T}_{\mathrm{seq}}$, the serializer $\mathcal{S}_{\tau}$ maps its target $Y_\tau$ to a canonical token sequence that is generated autoregressively and deterministically deserialized.

\textbf{Textual Answers and Sparse Geometric States.}
Some spatial QA tasks require only a small number of query-specific geometric quantities rather than a dense field. We serialize these quantities, such as depths at queried points and numerical spatial attributes, in the order specified by the instruction, followed by the final answer when applicable. Text-only QA directly uses the answer tokens. This design provides sparse geometric supervision while retaining standard autoregressive decoding.

\textbf{Structured Camera Poses.}
For structured outputs like camera poses, we use special tokens to represent them. We parameterize a camera pose $(R,\mathbf{t})$ as
\begin{equation}
    q
    =
    \operatorname{Quat}(R),
    \quad
    d
    =
    \frac{\mathbf{t}}{\|\mathbf{t}\|_2},
    \quad
    r
    =
    \|\mathbf{t}\|_2,
\label{eq:pose_parameterization}
\end{equation}
where $q$ is the rotation quaternion, and
$(d,r)$ denote the translation direction and magnitude,
respectively. Their scalar components are quantized with a resolution of $10^{-3}$, mapped to dedicated numerical tokens, and serialized in a fixed order.

\paragraph{Dense outputs as Image-Aligned Fields.}
For each task $\tau\in\mathcal{T}_{\mathrm{field}}$, the deterministic
transform $\Phi_\tau$ maps $Y_\tau$ to image-aligned fields. These fields follow the shape of RGB images while encoding geometric quantities. Before VAE encoding, each field is rescaled to the input range of the VAE.

\textbf{Depth.}
Given a depth map $D$, we normalize its values as
\begin{equation}
    \widehat{D}=1-\frac{D-d_{\min}}{d_{\max}-d_{\min}+\epsilon},
\label{eq:depth_encoding}
\end{equation}
where $d_{\min}$ and $d_{\max}$ denote the minimum and maximum depths in the image. Nearby and distant regions are mapped toward $1$ and $0$, respectively, and the resulting relative-depth representation is replicated across three channels.

\textbf{Point Maps.}
All point maps $\{P_i\}_{i=1}^{N}$ are expressed in a common coordinate system whose origin is the first camera. We normalize them using a center and scale shared across the sequence:
\begin{equation}
\begin{aligned}
    c
    &=
    \frac{1}{2N}
    \sum_{i=1}^{N}
    (
        \min_{p\in\Omega_i} P_i(p)
        +
        \max_{p\in\Omega_i} P_i(p)
    ),\\
    s
    &=
    \max_{i,\,p\in\Omega_i}
    \|P_i(p)-c\|_{\infty},
    \quad
    \widehat{P}_i
    =
    \frac{P_i-c}{s+\epsilon},
\end{aligned}
\label{eq:pointmap_encoding}
\end{equation}
where the minimum and maximum are computed element-wise over pixels. The three channels of $\widehat{P}_i$ encode the normalized Cartesian coordinates $(X,Y,Z)$. Sharing $c$ and $s$ across the sequence preserves relative geometry across views.

\textbf{Optical Flow.}
Given an optical flow field $(u,v)$, we normalize the displacements
by the image dimensions and apply a signed square-root transform:
\begin{equation}
\begin{aligned}
    \widehat{u}
    =
    \rho(\frac{u}{W}),
    \quad
    \widehat{v}
    =
    \rho(\frac{v}{H}),
    \quad
    \rho(x)
    =
    \operatorname{sgn}(x)\sqrt{|x|},
\end{aligned}
\label{eq:flow_encoding}
\end{equation}
where only the first two channels are used for decoding and the third channel encodes the flow magnitude. 

Additionally, we design an optical flow refinement procedure. At inference time, SPARGen refines optical flow through a predict–warp–predict procedure. Given the current estimate $F$, we align the second frame to the first as $I_2'(x)=I_2(x+F(x))$, and predict a residual flow $\Delta F$ between $I_1$ and $I_{2}'$. The estimated flow is given by $F+\Delta F$. 


\subsection{Training Objectives}
\label{sec:training_objectives}
SPARGen is jointly trained on a mixture of token-sequence and dense-field generation tasks.

\textbf{Sequence Generation Objective.}
For a task $\tau\in\mathcal{T}_{\mathrm{seq}}$, let
$\mathcal{S}_{\tau}(Y_\tau)=(s_1,\ldots,s_{L_\tau})$
denote its serialized target. We minimize the average cross-entropy over the target tokens:
\begin{equation}
    \mathcal{L}_{\mathrm{seq}}
    =
    -\frac{1}{L_\tau}
    \sum_{j=1}^{L_\tau}
    \log
    p_{\theta}(
        s_j
        \mid
        C_{\mathrm{seq}},s_{<j}
    ).
\label{eq:sequence_loss}
\end{equation}
Cross-entropy is evaluated only at target-token positions.

\textbf{Dense Field Generation Objective.}
For a task $\tau\in\mathcal{T}_{\mathrm{field}}$, we optimize the rectified-flow matching objective
\begin{equation}
    \mathcal{L}_{\mathrm{field}}
    =
    \mathbb{E}_{t,\epsilon}
    [
        \frac{1}{d_\tau}
        \|
            v_{\theta}(
                z_t,t;C_{\mathrm{field}}
            )
            -
            u_t
        \|_2^2
    ],
\label{eq:field_loss}
\end{equation}
where
$t\sim\mathcal{U}(0,1)$,
$\epsilon\sim\mathcal{N}(0,\mathbf{I})$,
and $d_\tau$ denotes the total number of scalar elements in the target latent representation. 

Each training example activates the objective associated with its target representation. The per-example loss is
\begin{equation}
    \mathcal{L}_{\tau}
    =
    \begin{cases}
        \lambda\mathcal{L}_{\mathrm{seq}},
        & \tau\in\mathcal{T}_{\mathrm{seq}},\\[1mm]
        \mathcal{L}_{\mathrm{field}},
        & \tau\in\mathcal{T}_{\mathrm{field}},
    \end{cases}
\label{eq:task_loss}
\end{equation}
where $\lambda$ controls the weight of the sequence generation objective. The overall training objective is $\mathcal{L}=\mathbb{E}_{(\mathcal{I},p,Y_\tau,\tau)\sim\mathcal{D}}
    [
        \mathcal{L}_{\tau}
    ]$, where $\mathcal{D}$ denotes the training distribution.

\begin{table*}[!htb]
  \centering

  \small
  \setlength{\tabcolsep}{5pt}
  \renewcommand{\arraystretch}{1.15}

  \resizebox{\textwidth}{!}{%
  \begin{tabular}{lccccccccccc}
    \toprule
    \multirow{3}{*}{\textbf{Model}}
      & \multicolumn{4}{c}{\textbf{Depth Estimation}}
      & \multicolumn{4}{c}{\textbf{Point Map Estimation}}
      & \multicolumn{3}{c}{\textbf{Camera Pose Estimation}} \\
    \cmidrule(lr){2-5}
    \cmidrule(lr){6-9}
    \cmidrule(lr){10-12}

      & \multicolumn{2}{c}{\textbf{Sintel}}
      & \multicolumn{2}{c}{\textbf{NYU-v2}}
      & \multicolumn{2}{c}{\textbf{7Scenes}}
      & \multicolumn{2}{c}{\textbf{ETH3D}}
      & \multicolumn{3}{c}{\textbf{Co3Dv2}} \\
    \cmidrule(lr){2-3}
    \cmidrule(lr){4-5}
    \cmidrule(lr){6-7}
    \cmidrule(lr){8-9}
    \cmidrule(lr){10-12}

      & AbsRel$\downarrow$ & $\delta_1\uparrow$
      & AbsRel$\downarrow$ & $\delta_1\uparrow$
      & Acc.$\downarrow$   & Comp.$\downarrow$
      & Acc.$\downarrow$   & Comp.$\downarrow$
      & RRA@30$\uparrow$   & RTA@30$\uparrow$
      & AUC@30$\uparrow$ \\
    \midrule

    \multicolumn{12}{l}{\textit{Specialized Geometry Models}} \\
    FLARE~\cite{zhang2025flare}
      & 0.409 & 0.438
      & 0.164 & 0.740
      & 0.035 & 0.043
      & 0.470 & 0.667
      & 97.08 & 94.31 & 77.99 \\
    DUSt3R~\cite{wang2024dust3r}
      & 0.362 & 0.557
      & 0.134 & 0.833
      & 0.026 & 0.035
      & 0.360 & 0.401
      & 98.34 & 94.09 & 79.33 \\
    VGGT~\cite{wang2025vggt}
      & \textbf{0.265} & \textbf{0.676}
      & \textbf{0.065} & \textbf{0.938}
      & \textbf{0.022} & \textbf{0.032}
      & \textbf{0.311} & \textbf{0.372}
      & \textbf{98.79} & \textbf{96.89} & \textbf{89.78} \\
    \midrule

    \multicolumn{12}{l}{\textit{Spatial Unified Models}} \\
    G$^2$VLM~\cite{hu2026g}
      & 0.257 & 0.674
      & 0.079 & \textbf{0.935}
      & 0.062 & 0.031
      & 0.539 & \textbf{0.355}
      & 96.69 & 92.22 & 56.85 \\
    \rowcolor{rowgray} \textbf{SPARGen (Ours) }
      & \textbf{0.235} & \textbf{0.725}
      & \textbf{0.071} & \textbf{0.935}
      & \textbf{0.034} & \textbf{0.028}
      & \textbf{0.393} & 0.445
      & \textbf{96.84} & \textbf{94.33} & \textbf{74.32} \\
    \bottomrule
  \end{tabular}%
  }
  \caption{Comparison across depth estimation, point map estimation, and camera pose estimation benchmarks. \textbf{Bold} indicates the best performance within each model group.}
  \label{tab:vg}
\end{table*}

\begin{table*}[htbp]
  \centering

  \small
  \setlength{\tabcolsep}{4.5pt}
  \renewcommand{\arraystretch}{1.15}

  \resizebox{\textwidth}{!}{%
  \begin{tabular}{lccccccccccccccc}
    \toprule
 SPARGen   \multirow{2}{*}{\textbf{Model}}
      & \multicolumn{4}{c}{\textbf{MindCube}}
      & \multicolumn{3}{c}{\textbf{OmniSpatial}}
      & \multicolumn{4}{c}{\textbf{OST}}
      & \multicolumn{4}{c}{\textbf{SPAR}} \\
    \cmidrule(lr){2-5}
    \cmidrule(lr){6-8}
    \cmidrule(lr){9-12}
    \cmidrule(lr){13-16}

      & Avg. & Rotation & Among & Around
      & Avg. & SI & PT
      & Avg. & A. State & A. Info & AO.
      & Avg. & Low & Medium & High \\
    \midrule

    \multicolumn{16}{l}{\textit{Proprietary Models}} \\
    GPT-4o~\cite{hurst2024gpt}
      & 41.52 & 34.50 & 41.67 & 46.80
      & 42.39 & 48.67 & 39.04
      & 51.42 & 40.93 & 68.38 & 39.25
      & 37.88 & 35.36 & 28.40 & 43.27 \\
    Claude Sonnet 4.6~\cite{anthropic2025claude4}
      & 44.19 & 42.50 & 43.00 & 48.40
      & 53.89 & 67.00 & 46.88
      & 42.18 & 35.84 & 63.30 & 29.35
      & 37.55 & 36.63 & 35.33 & 39.20 \\
    \midrule

    \multicolumn{16}{l}{\textit{Open-Source Models}} \\
    Qwen2.5-VL-7B~\cite{Qwen2.5-VL}
      & 36.00 & 37.50 & 32.33 & 43.60
      & 45.99 & 54.67 & 41.35
      & 40.08 & 38.48 & 50.20 & 31.27
      & 34.84 & 26.36 & 36.09 & 41.97 \\
    Qwen2.5-VL-72B~\cite{Qwen2.5-VL}
      & 41.71 & 40.50 & 40.67 & 45.20
      & \underline{52.03} & \textbf{61.33} & \underline{47.06}
      & 49.23 & 39.05 & 65.78 & 37.37
      & 39.34 & 34.29 & \underline{36.84} & 44.80 \\
    LLaVA-Video-7B~\cite{zhang2024videoinstructiontuningsynthetic}
      & 43.24 & 36.50 & 43.50 & 48.00
      & 49.71 & \underline{55.33} & 46.70
      & 44.54 & 32.31 & 62.25 & 32.29 
      & 34.09 & 24.99 & 31.77 & 43.10 \\
    LLaVA-OneVision-7B~\cite{li2024llava}
      & 40.86 & 33.50 & 37.17 & 55.60
      & 46.92 & 52.67 & 43.85
      & 35.87 & 24.97 & 44.85 & 31.22
      & 29.47 & 19.88 & 30.49 & 37.66 \\
    Bagel-7B~\cite{deng2025emerging}
      & 41.20 & 36.50 & 39.50 & 52.40 
      & 45.29 & 49.33 & 43.14 
      & 32.73 & \underline{42.87} & 26.49 & 35.11 
      & 39.51 & 34.59 & 36.19 & 45.15 \\
    \midrule

    \multicolumn{16}{l}{\textit{Spatial Expert Models}} \\
    VLM3R-7B~\cite{fan2026vlm}
      & 37.90 & 36.50 & 42.50 & 28.00
      & 46.69 & 49.33 & 45.28
      & \underline{51.03} & 38.32 & \underline{68.30} & \underline{39.35}
      & \underline{41.89} & 36.36 & 28.38 & \underline{51.92} \\
    Spatial-MLLM-7B~\cite{ma2025spatialllm}
      & \underline{66.19} & \underline{41.00} & \underline{66.83} & \underline{74.80}
      & 45.30 & 47.00 & 44.39
      & 39.90 & 32.28 & 46.80 & 36.08
      & 33.92 & 23.90 & 33.74 & 42.95 \\
    \midrule

    \multicolumn{16}{l}{\textit{Spatial Unified Models}} \\
    G$^2$VLM-2B~\cite{hu2026g}
      & 28.48 & 25.50 & 30.17 & 26.80
      & 42.62 & 44.33 & 41.71
      & 28.80 & 27.13 & 35.43 & 23.24
      & 38.73 & \textbf{51.24} & 28.82 & 31.27 \\
    \rowcolor{rowgray} \textbf{SPARGen-7B (Ours)}
      & \textbf{76.04} & \textbf{63.50} & \textbf{79.16} & \textbf{78.60}
      & \textbf{54.00} & \underline{55.33} & \textbf{53.29}
      & \textbf{56.02} & \textbf{47.10} & \textbf{72.30} & \textbf{43.98}
      & \textbf{66.60} & \underline{49.35} & \textbf{74.04} & \textbf{79.25} \\
    \bottomrule
  \end{tabular}%
  }
  \caption{Results on spatial-reasoning benchmarks. \textbf{Bold} and \underline{underlined} values indicate the best and second-best results among non-proprietary models.}
  \label{tab:sr}
\end{table*}

\begin{table*}[!ht]
  \centering

  \begin{minipage}[t]{0.39\textwidth}
    \centering
    \setlength{\tabcolsep}{5pt}
    \resizebox{0.87\textwidth}{!}{
    \begin{tabular}[t]{@{}lcc@{}}
      \toprule
      \multirow{2}{*}{\textbf{Model}}
        & \multicolumn{2}{c}{\textbf{KITTI}} \\
      \cmidrule(lr){2-3}
        & EPE$\downarrow$
        & F1-all$\downarrow$ \\
      \midrule
      RAFT~\cite{teed2020raft}
        & 5.03 & 17.45 \\
      GMFlow~\cite{xu2022gmflow}
        & 7.77 & 23.40 \\
      FlowFormer~\cite{huang2022flowformer}
        & 4.10 & 14.51 \\
      SPARGen w/o refinement
        & 5.26
        & 21.82 \\
      \rowcolor{rowgray} \textbf{SPARGen (Ours)}
        & \textbf{4.09}
        & \textbf{13.34} \\
      \bottomrule
    \end{tabular}
    }
    \caption{Zero-shot optical flow results on KITTI. \textbf{Bold} values indicate the best performance.}
    \label{tab:kitti_optical_flow}
  \end{minipage}
  \hfill
  \begin{minipage}[t]{0.59\textwidth}
    \centering
    \setlength{\tabcolsep}{2.5pt}
    \resizebox{\textwidth}{!}{
    \begin{tabular}[t]{@{}lcccccccc@{}}
      \toprule
      \multirow{2}{*}{\textbf{Model}}
        & \multicolumn{2}{c}{\textbf{7Scenes}}
        & \multicolumn{2}{c}{\textbf{KITTI}}
        & \multicolumn{4}{c}{\textbf{SPAR}} \\
      \cmidrule(lr){2-3}
      \cmidrule(lr){4-5}
      \cmidrule(lr){6-9}

        & Acc.$\downarrow$
        & Comp.$\downarrow$
        & EPE$\downarrow$
        & F1-all$\downarrow$
        & Avg.$\uparrow$
        & Low$\uparrow$
        & Med.$\uparrow$
        & High$\uparrow$ \\
      \midrule

      w/o Geometry
        & - & -
        & 4.29 & 14.73
        & 62.41 & 45.97 & 66.49 & 75.59 \\

      w/o Flow
        & 0.040 & 0.038
        & - & -
        & 65.92 & 48.41 & 73.52 & 78.75  \\

      w/o Reasoning
        & 0.037 & 0.031
        & \textbf{4.06} & \textbf{13.01}
        & - & - & - & - \\

      \rowcolor{rowgray} \textbf{SPARGen (Ours)}
        & \textbf{0.034} & \textbf{0.028}
        & 4.09 & 13.34
        & \textbf{66.60} & \textbf{49.35} & \textbf{74.04} & \textbf{79.25} \\

      \bottomrule
    \end{tabular}
    }
    \caption{Ablation of geometry, optical flow, and spatial-reasoning supervision. \textbf{Bold} values indicate the best performance.}
    \label{tab:ablation}
  \end{minipage}
  \vspace{-0.4cm}
\end{table*}

\begin{figure*}[!htbp]
    \centering
    \includegraphics[width=1.0\linewidth]{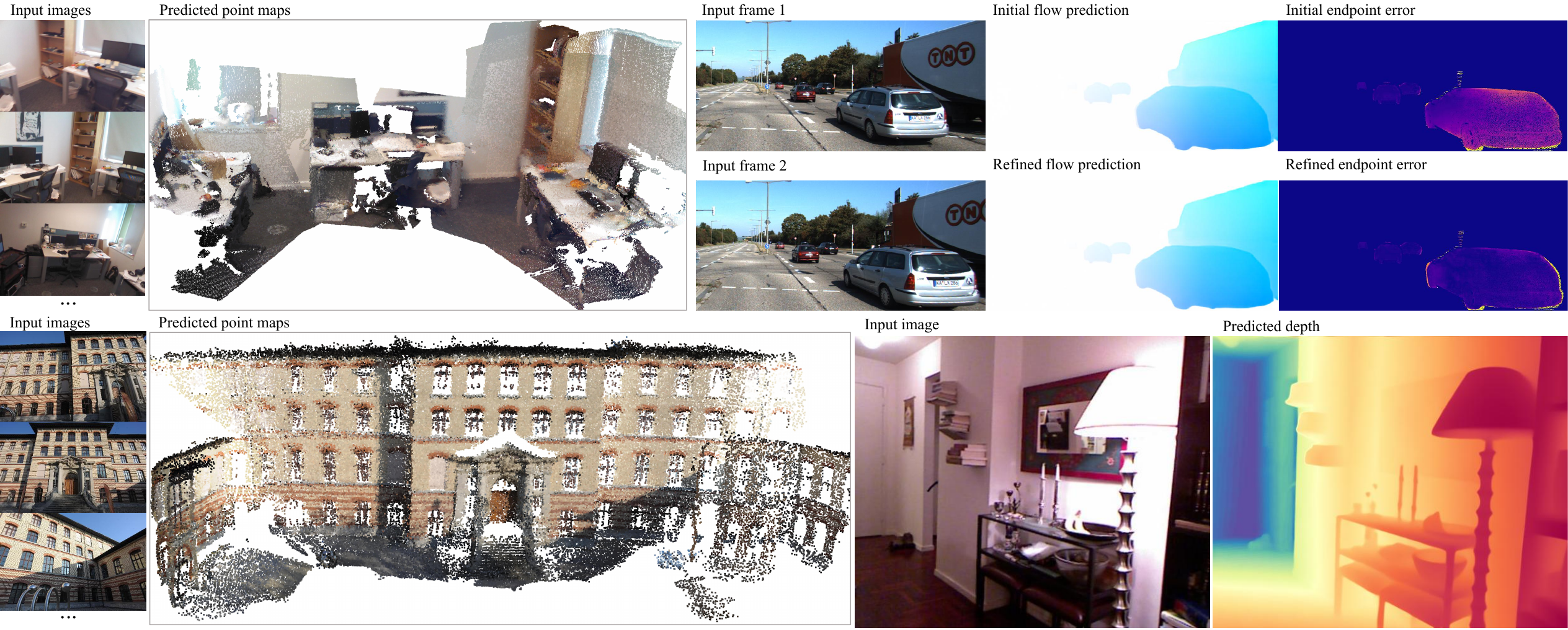}
    \caption{Qualitative results of SPARGen on point map reconstruction, optical flow estimation, and depth estimation. For optical flow, we show the predictions and error maps before and after refinement.}
    \label{fig:vis}
    \vspace{-0.5cm} 
\end{figure*}

\section{Experiments}
\subsection{Experimental Settings}

\textbf{Training Datasets.}
We train SPARGen using three groups of supervision: spatial understanding, visual geometry, and optical flow.
1) For spatial reasoning, following G$^2$VLM~\cite{hu2026g}, we use the official training splits of MindCube, OmniSpatial, OST-Bench, SPAR-7M and general VQA dataset LLaVA-OneVision.
2) For visual geometry, including relative-depth estimation, camera pose prediction, and multi-view point map reconstruction, we aggregate training data from ASE, BlendedMVS, CO3D, DeMoN, DL3DV, Hypersim, IRS, MegaSynth, MVS-Synth, Objaverse, OmniObject3D, ScanNet v2, ScanNet++, SceneNet RGB-D, Taskonomy, and WildRGB-D. Depending on the available annotations, each sample may supervise one or more visual-geometry tasks. Although several datasets provide metric geometric annotations, we normalize the depth and point map targets and train the model in a relative-scale coordinate system. For samples with sparse or incomplete geometric annotations, we additionally use MoGe~\cite{wang2025moge} to generate dense, image-aligned geometric pseudo-labels for training the dense-field pathway.
3) For optical flow estimation, we use TartanAir, AutoFlow, FlyingChairs, FlyingChairs2, FlyingThings3D, Monkaa, Kubric-4D, ParallelDomain-4D, and Spring.

\textbf{Benchmarks.} We follow the evaluation protocol of G$^2$VLM~\cite{hu2026g}, the most closely related baseline. 
1) For visual-geometry evaluation, we use Sintel~\cite{bozic2021transformerfusion} and NYU-v2~\cite{silberman2012indoor} for depth estimation, 7Scenes~\cite{shotton2013scene} and ETH3D~\cite{schops2017multi} for 3D reconstruction, and CO3D v2~\cite{reizenstein2021common} for camera pose estimation. 
2) For spatial reasoning, we evaluate on the official test sets of MindCube Tiny~\cite{yin2025spatial}, OmniSpatial~\cite{jia2025omnispatial}, OST-Bench~\cite{lin2026ost}, and SPAR-Bench~\cite{zhang2026flatland}. 
3) For optical flow estimation, we evaluate zero-shot transfer on the KITTI training set~\cite{Geiger2013IJRR}, without fine-tuning on the benchmark.

\textbf{Baselines and Metrics.} We organize our comparisons into three task groups. 1) For visual geometry, we compare SPARGen with the specialized models DUSt3R~\cite{wang2024dust3r}, FLARE~\cite{zhang2025flare}, and VGGT~\cite{wang2025vggt}, as well as the unified model G$^2$VLM~\cite{hu2026g}. We report AbsRel and $\delta_1$ for depth estimation; reconstruction accuracy error (Acc.) and completeness error (Comp.) for 3D reconstruction; and relative rotation accuracy (RRA), relative translation accuracy (RTA), and area under the curve (AUC) for camera pose estimation.
2) For optical flow, we compare with RAFT~\cite{teed2020raft}, GMFlow~\cite{xu2022gmflow}, and FlowFormer~\cite{huang2022flowformer}, reporting end-point error (EPE) and F1-all.
3) For spatial reasoning, we compare with the proprietary models GPT-4o~\cite{hurst2024gpt} and Claude Sonnet 4.6~\cite{anthropic2025claude4}; general-purpose vision-language models Qwen2.5-VL-7B/72B~\cite{Qwen2.5-VL}, LLaVA-Video~\cite{zhang2024videoinstructiontuningsynthetic}, and LLaVA-OneVision~\cite{li2024llava}; the spatial-specialist models Spatial-MLLM~\cite{ma2025spatialllm} and VLM3R-7B~\cite{fan2026vlm}; and the unified model G$^2$VLM~\cite{hu2026g}. We report answer accuracy on each benchmark. 

\textbf{Implementation Details.}
We initialize SPARGen from the pretrained Bagel weights and jointly optimize the token-sequence and dense-field generation objectives. The VAE encoder and decoder are frozen, while all other model parameters are fine-tuned for 100K iterations using AdamW on 64 NVIDIA H100 GPUs. For depth estimation, point map reconstruction, and optical flow estimation, the maximum ViT input resolutions are set to $518$, $448$, and $980$, respectively, and the corresponding maximum VAE input resolutions are $1024$, $512$, and $1024$. We set the sequence-loss weight in Eq.~\ref{eq:task_loss} to $\lambda=0.25$ and use a learning rate of $2.5\times10^{-5}$. SPARGen predicts normalized depth and point maps, and therefore, like VGGT~\cite{wang2025vggt} and G$^2$VLM~\cite{hu2026g}, cannot recover metric scale. We follow their standard scale-aligned evaluation protocol.

More details can be found in the appendix.

\subsection{Comparisons with Prior Work}
We compare SPARGen with specialized models and spatially unified models across visual geometry, spatial reasoning, and optical flow benchmarks. SPARGen remains competitive with specialized geometry and flow models, and achieves the strongest spatial-reasoning performance among the compared models. These results demonstrate that spatial capabilities can be supported within a native multimodal model. 

\textbf{Visual Geometry.}
Table~\ref{tab:vg} reports results on depth estimation, point map reconstruction, and camera pose estimation. Compared with the most relevant spatially unified baseline, G$^2$VLM, SPARGen performs better on most metrics. In particular, it improves both depth metrics on Sintel, reduces AbsRel while matching $\delta_1$ on NYU-v2, and achieves better reconstruction results on 7Scenes. It also improves camera pose estimation across all three CO3D v2 metrics. Although specialized geometry models such as VGGT retain an advantage on several reconstruction metrics, SPARGen substantially narrows the gap with the native unified model.

\textbf{Spatial Understanding and Reasoning.}
As shown in Table~\ref{tab:sr}, SPARGen achieves the highest average score on all four spatial-reasoning benchmarks and ranks first in 13 of the 15 reported categories among the compared non-proprietary models. Relative to the strongest competing result on each benchmark, SPARGen improves the average score by 9.85 points on MindCube, 1.97 points on OmniSpatial, 4.99 points on OST, and 24.71 points on SPAR. The gains are particularly pronounced on the medium and high subsets of SPAR. While SPARGen benefits from a 7B foundation backbone, its gains cannot be attributed solely to model scale, as it also outperforms substantially larger open-source baselines such as Qwen2.5-VL-72B.

\textbf{Optical Flow Estimation.}
Table~\ref{tab:kitti_optical_flow} reports zero-shot optical flow results on KITTI. SPARGen achieves an EPE of 4.09 and an F1-all score of 13.34, outperforming the compared methods on both metrics. These results show that SPARGen can learn effective dense correspondence estimation through its native generation pathway. 

\textbf{Qualitative Results.}
Figure~\ref{fig:vis} presents representative predictions for 3D reconstruction, optical flow estimation, and depth estimation. For 3D reconstruction, SPARGen recovers coherent global layouts in both indoor and outdoor scenes. Its optical flow predictions capture the dominant motion of foreground vehicles while producing relatively consistent estimates over static background regions. The refinement stage further corrects residual misalignments in the initial predictions. For depth estimation, SPARGen recovers the overall near-to-far structure of the scene and preserves the boundaries of major objects. 

Overall, SPARGen improves over the existing unified baseline on visual geometry, achieves the strongest spatial-reasoning results, and remains competitive with specialized optical flow methods, all within a shared instruction-conditioned generative framework.

\subsection{Ablation Studies and Analysis}
To examine whether the results are consistent with complementary effects among different supervision types, we train variants with one supervision category removed at a time.

\textbf{(1) Removing Geometry Supervision.}
We remove geometry supervision. This ablation examines whether 3D supervision provides transferable structural information for dense correspondence and spatial reasoning.

\textbf{(2) Removing Optical Flow Supervision.}
We remove optical flow supervision. This setting evaluates whether dense correspondence learning contributes to multi-view reconstruction and spatial reasoning.

\textbf{(3) Removing Reasoning Supervision.}
We remove spatial-reasoning and question-answering supervision. This ablation examines whether high-level semantic objectives can improve dense geometric perception.

\textbf{Results \& Analysis.}
Table~\ref{tab:ablation} presents the results. 
(1) Removing geometry supervision degrades both optical flow estimation and spatial reasoning, suggesting that explicit 3D supervision may provide structural information useful for dense correspondence and language-based reasoning. 
(2) Removing optical flow supervision increases the reconstruction accuracy and completeness errors on 7Scenes and reduces the average SPAR score, consistent with dense correspondence supervision benefiting cross-view consistency and spatial reasoning. 
(3) Removing reasoning supervision slightly degrades reconstruction, suggesting a possible benefit of high-level semantic supervision for geometric representation learning. However, optical flow performance improves slightly in the absence of reasoning supervision. We attribute this to a mild capacity competition in multi-task learning: while semantic reasoning aids static 3D structure, the autoregressive token generation slightly competes for the MoT backbone's capacity against dynamic dense-field prediction.

\section{Conclusion}
We presented SPARGen, a unified multimodal framework that formulates spatial perception and reasoning as instruction-conditioned generation. SPARGen serializes compact structured and linguistic outputs as token sequences and represents dense geometric and correspondence fields as image-aligned outputs. It thereby uses the native autoregressive and rectified-flow pathways of a shared MoT backbone without external modules. Across visual-geometry, optical flow, and spatial-reasoning benchmarks, SPARGen remains competitive with specialized perception methods, and achieves strong spatial-reasoning performance. Our ablations provide preliminary evidence consistent with complementary effects among geometry, correspondence, and reasoning supervision. These findings demonstrate the feasibility of using native multimodal generation as a shared interface for spatial perception and reasoning tasks.

\textbf{Limitations.}
While leveraging a frozen VAE allows SPARGen to reuse pretrained multimodal generative pathways, VAE spatial compression inherently poses a bottleneck for geometric edges and high-precision physical quantities.

\bibliography{aaai2027}


\end{document}